\documentclass{article}

\usepackage{arxiv}

\usepackage[utf8]{inputenc} 
\usepackage[T1]{fontenc}    
\usepackage{hyperref}       
\usepackage{url}            
\usepackage{booktabs}       
\usepackage{amsfonts}       
\usepackage{nicefrac}       
\usepackage{microtype}      
\usepackage{lipsum}
\usepackage{graphicx}
\usepackage{color}
\usepackage{tablefootnote}
\usepackage{multirow}
\graphicspath{ {./images/} }

\title{American Hate Crime Trends Prediction with Event Extraction}

\author{
 Songqiao Han\textsuperscript{$\ast$} 
   \And
 Hailiang Huang\textsuperscript{$\ast$} 
  \And
 Jiangwei Liu\textsuperscript{$\ast$} 
  \And
 Shengsheng Xiao\thanks{All authors contribute equally.} \\
 \And
 \\
 School of Information Management and Engineering,\\
 Shanghai University of Finance and Economics, Shanghai 200433, China \\
}

\begin{document}
\maketitle

\begin{abstract}
Social media platforms may provide potential space for discourses that contain hate speech, and even worse, can act as a propagation mechanism for hate crimes. The FBI's Uniform Crime Reporting (UCR) Program collects hate crime data and releases statistic report yearly. These statistics provide information in determining national hate crime trends. The statistics can also provide valuable holistic and strategic insight for law enforcement agencies or justify lawmakers for specific legislation. However, the reports are mostly released next year and lag behind many immediate needs. Recent research mainly focuses on hate speech detection in social media text or empirical studies on the impact of a confirmed crime. This paper proposes a framework that first utilizes text mining techniques to extract hate crime events from New York Times news, then uses the results to facilitate predicting American national-level and state-level hate crime trends. Experimental results show that our method can significantly enhance the prediction performance compared with time series or regression methods without event-related factors. Our framework broadens the methods of national-level and state-level hate crime trends prediction.
\end{abstract}


\section{Introduction}
People increasingly communicate opinions and share information through social network platforms such as online forums, blogs, Facebook, etc. Although many of the interactions among the consumer-generated content (CGC) are positive conversations, the proportion of full content containing abusive and hateful language is significantly increasing. Concentrated outbreaks of hate language on certain activities can lead to hate crimes. For example, due to Trump's inflammatory rhetoric through the political campaign, relative hate crimes such as anti-muslim or anti-immigrant hate crimes significantly rise compared to that several years before \footnote[1]{'Trump effect' led to hate crime surge, report finds: https://www.bbc.com/news/world-us-canada-38149406.}. 

The FBI's UCR Program describes hate crime as "a committed criminal offense that is motivated, in whole or in part, by the offender's bias(es) against a race, religion, disability, sexual orientation, ethnicity, gender, or gender identity" \footnote{What is a hate crime? Is available at https://www.fbi.gov/services/cjis/ucr/hate-crime.}. We review the literature on hate crime study and find two prevailing genres: econometric and computer science genres. (1) The econometric genre focuses on empirical studies on the correlation between hate crimes and their affecting factors. For example, \cite{CramerWright-394} studied the relationship between depression, impulsivity, or post-traumatic stress and sexual orientation minority adults' victimization–suicide risk. The results set the backdrop for sexual orientation minority health risk and suicide prevention policy research. However, this kind of work mainly focuses on one category of hate crime analysis. There is limited research on national or local hate crime trend prediction. (2) The computer science genre focuses on hate speech detection, which assists in predicting and preventing cybercrimes. Keywords-based approaches and deep-learning-based approaches are two mainstream methods for hate speech detection. For example, \cite{GitariZuping-395} created a lexicon related to hate speech and used semantic features to detect the presence of hate speech in web discourses. \cite{MozafariFarahbakhsh-396} introduced a novel neural network based on the pre-trained language model BERT (Bidirectional Encoder Representations from Transformers) to identify hateful contents. However, there is limited research utilizing computer science technologies to predict national or local hate crime trends.

Unlike the above research, we first utilize the deep-learning method to extract hate crime events from New York Times news reports. Then we construct three event-related factors and add them into our econometric regression model to predict the FBI hate crime trends. Our study is inspired by previous work \cite{Mostafazadeh-DavaniYeh-397} that used event extraction technologies to report local hate crimes not covered by the FBI’s hate crime reports. We adopt this event extraction approach, train the event extraction model on Patch corpus, and then apply the trained model to identify incidents of hate crime reported in the New York Times corpus. We build a time series model and a series of regression models as our base models. After adding event-related factors into our regression model, we estimate the parameters by Maximum Likelihood Estimation (MLE) and predict the FBI hate crime trends. The experimental results demonstrate that event-related factors can significantly enhance the model performance. Our framework and findings can provide valuable holistic and strategic insight for law enforcement agencies and justify specific legislation in advance.

The remainder of this paper is organized as follows. We review related work in section 2 and introduce the data and processing in section 3. We propose our framework and introduce its modules in section 4. Section 5 reports experimental results and discusses the shortcomings and future work, followed by the conclusion in Section 6.

\section{Related Work}

\subsection{Hate Speech Detection}
To prohibit hate speech, most social media platforms have established many rules to detect whether a post or reply contains hate speeches \cite{SilvaMondal-398}. Only automatic approaches can be qualified for this task due to massive posts or replies. We review the research on automatic hate speech detection and classify it into two main categories: keyword-based and machine learning approaches. 

Keyword-based approaches typically maintain a dictionary or database of hate speech ontologies and their related expressions, then utilize this dictionary to identify whether a post contains hateful keywords \cite{GitariZuping-395}. They are the most straightforward and efficient unsupervised methods to deal with this problem. However, they still have many limitations. For example, they cannot identify nuanced expressions (e.g., figurative expression) or contents containing hate speech but not using specific keywords in the maintained dictionary (e.g., abbreviation).

Machine learning approaches are currently the most popular methods in automatic hate speech detection. They usually treat hate speech detection as a text classifying task. Once the corpus is annotated by labeling whether a content contains hate speech or not, the standard text classifying machine learning methods can be adapted to hate speech detection. For example, \cite{MozafariFarahbakhsh-396} introduced a novel transfer learning approach based on an existing pre-trained language model BERT to detect hateful speech in social media content automatically.

\subsection{Event Extraction}
Event Extraction (EE) is an advanced form of Information Extraction (IE) that handles textual content or relations between entities, often performed after executing a series of initial NLP steps \cite{HogenboomFrasincar-53}. Event extraction from news articles is a commonly required prerequisite for various tasks, such as article summarization, article clustering, or news aggregation \cite{HamborgBreitinger-51}. From the view of technical implementation maps in closed domain event extraction, event extraction approaches include pattern matching methods and machine-learning-based methods (including deep-learning methods).

Pattern matching methods usually predefined patterns or schemas. Its advantages involve needing less training data, while its disadvantages involve deﬁning and maintaining patterns. For example, \cite{HamborgBreitinger-51} improved the universal EE model (Giveme5W1H) to detect main events in an article, which used syntactic and domain-specific rules to extract the relevant phrases to answer 5W1H questions automatically: who did what, when, where, why and how. \cite{XuLiu-52} proposed a hybrid collaborative filtering method for event recommendation in event-based social networks.

The advantage of machine learning or deep learning methods can automatically and effectively extract the significant features in the text. For instance, \cite{Mostafazadeh-DavaniYeh-397} trained a deep learning model to perform event extraction. They detected hate crimes on hyper-local news articles, analyzed the hate crime trend, and compared the results with the FBI reports. The results provided a lower-bound estimation on the occurrence frequency of hate crimes, especially in the cities not covered in FBI reports. In our study, we adopt this approach to extract hate crime events.

\subsection{Empirical Study on Hate Crimes }
Many empirical studies focus on the effectiveness of certain hate crimes or the relation between a hate crime and a specific event \cite{ReliaLi-402}. For example, \cite{HerekGillis-403} studied the psychological sequelae of hate-crime victimization among lesbian, gay, and bisexual adults. The result showed that compared with other recent crime victims, lesbian and gay hate-crime survivors manifested significantly more depression, anger, anxiety, and post-traumatic stress symptoms. The findings help highlight the importance of recognizing hate-crime survivors' special needs in clinical settings and public policies. Another recent example is about how the Covid-19 Pandemic sparked racial animus in the United States. \cite{LuSheng-404} collected Google searches and Twitter posts that contained anti-Asian racial hate speech to estimate the effect of the Covid-19 pandemic on racial animus. The results suggest that racial animus can be effectively curbed by deemphasizing the connection between the disease and a particular racial group.

\section{Data and Data Processing}
\subsection{Patch Hate Crime Data}
We use Patch Hate Crime Dataset \footnote{Patch Hate Crime Dataset is availible at https://github.com/aiida-/HateCrime} \cite{Mostafazadeh-DavaniYeh-397} to train the event extraction model. The dataset was original introduced by Mostafazadeh Davani and his colleagues. They collected hyper-local news from the Patch \footnote{https://www.patch.com} website and annotated a subset of this corpus. The annotated labels consisted of two aspects: one indicated whether the news was a hate crime event and, if it was, which kind of the hate crime attributes it belonged to. The corpus contains 5171 samples with 1979 positive samples and 3192 negative samples. 

\subsection{New York Times News Data}
The New York Times provides API \footnote{https://developer.nytimes.com/apis} to get its historical news reports month by month, returning titles, abstracts, keywords, URL, publish date, etc. However, the results do not contain news content. We utilize web crawler technology to collect the news contents through the corresponding URL from January 2007 to December 2020. Finally, there are 165913 news reports left after removing irrelevant categories or news with too short content. We grouped the news reports by quarter, counted the total reports number in each group, and lastly named this time series $news\_num$, with an average of 2962 reports per quarter. The previously trained hate crime model is performed on this corpus to detect whether a piece of news reports a hate crime event. We count the predicted result list and name this time series $event\_detected\_num$.

\subsection{FBI Hate Crime Data}
\begin{figure}[htbp]
	\centering
	\includegraphics[width= 5.5 in]{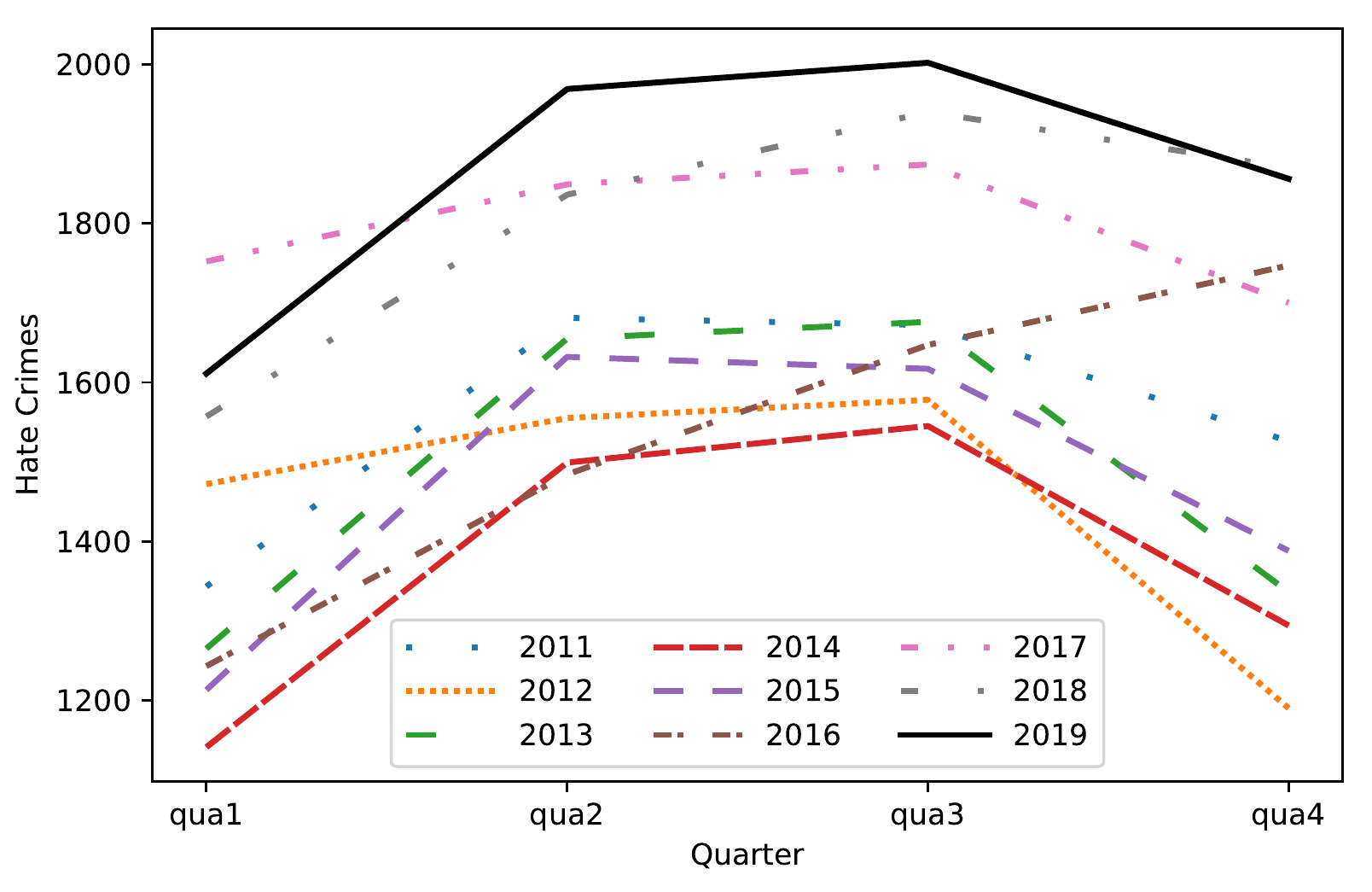}	
	\caption{Quarterly hate crimes.}
	\label{fig1}	
\end{figure}

We collect hate crimes data reported to the FBI in the United States from 2007 to 2019. The statistics \footnote{The statistics reports are available at https://www.fbi.gov/services/cjis/ucr/publications\#Hate-Crime\%20Statistics} reports hate crime incidents per bias motivation and quarter by state federal and agency. It also reports the population in the involved state federal or agency. It may be helpful to look not just at annual hate crime numbers but instead quarterly statistics. We plot the quarterly trends in Figure \ref{fig1} by breaking down the FBI hate crime data by quarter. 

\begin{figure}[htbp]
	\centering
	\includegraphics[width= 5.5 in]{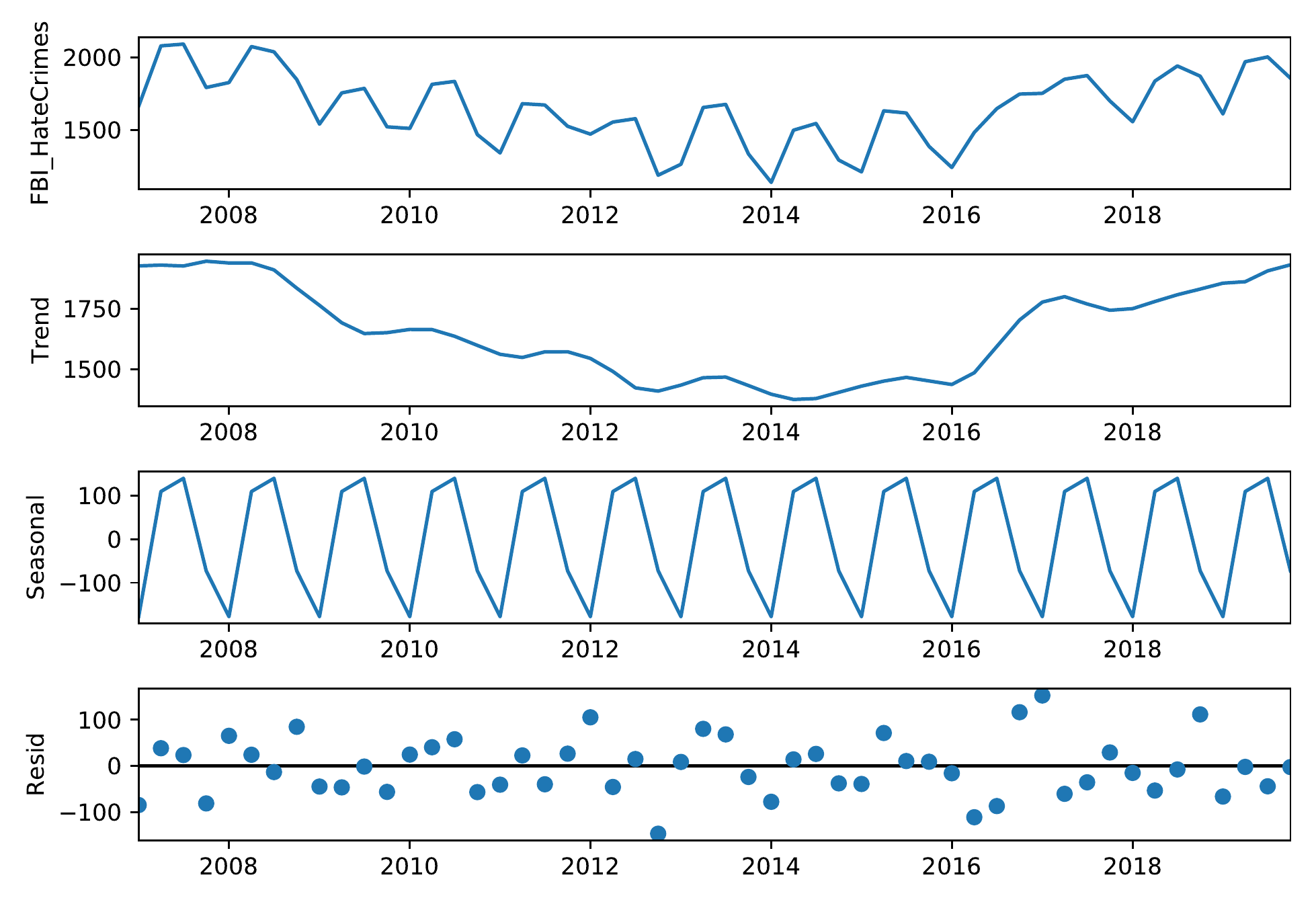}	
	\caption{Decompose quarterly hate crime time series into trend, seasonal, and irregular components.}
	\label{fig2}	
\end{figure}

Cyclical patterns emerging in Figure 1 mean that the quarterly hate crime time series may have a seasonal trend. Augmented Dickey-Fuller Test (ADF test) shows it has a unit root that means it is a nonstationary time series. By decomposing this time series into trend, seasonal, and irregular components using the moving average method, we get its corresponding components and plot them in Figure \ref{fig2}. Ljung-Box Tests results ($P\_value=0.005$) on irregular components show it has been white noises and thus demonstrates the decomposing adequacy. Next, we would build models on trended data. We estimate the parameters using the data between 2007 quarter 1 and 2018 quarter 4 and forecast the hate crimes between 2019 quarter 1 and 2019 quarter 4.

\section{Our Approach}
\begin{figure}[htbp]
	\centering
	\includegraphics[width= 5.5 in]{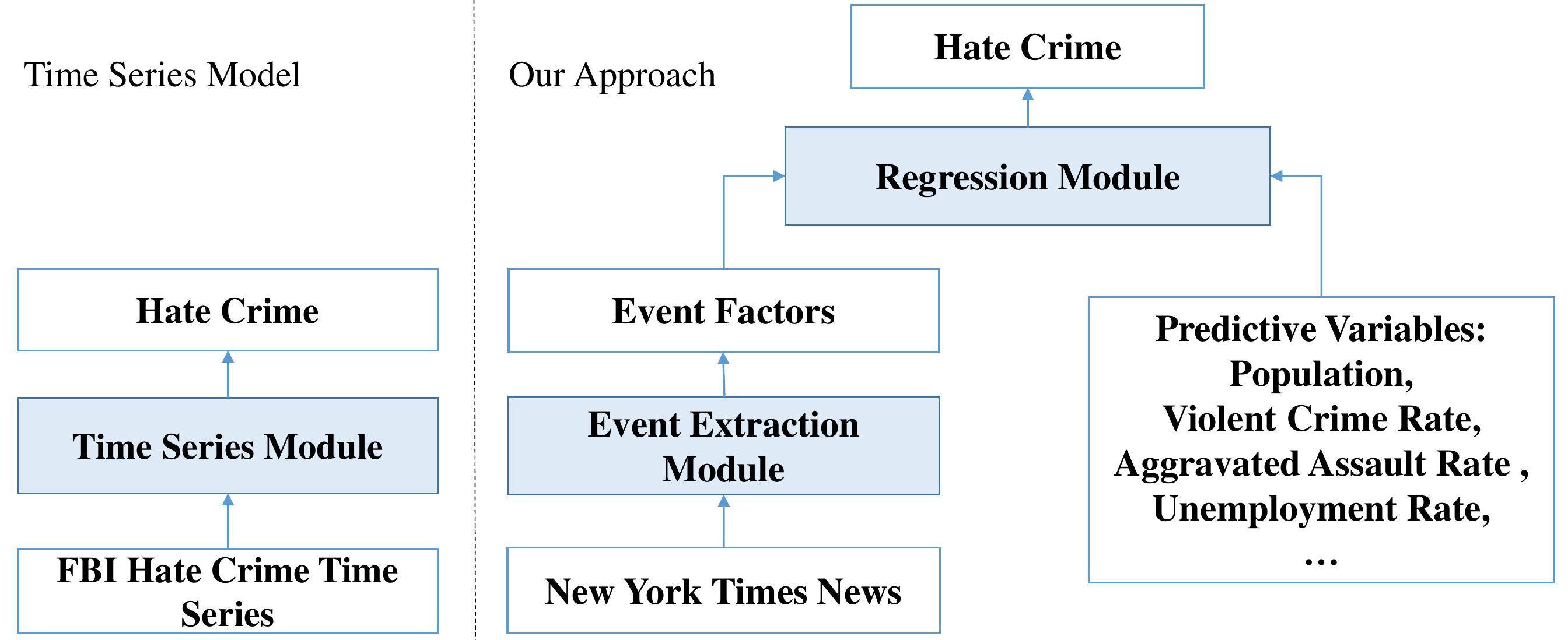}	
	\caption{Hate crime prediction framework.}
	\label{fig3}	
\end{figure}

In this section, we introduce our hate crime prediction framework. As illustrated in Figure \ref{fig3}, there are two kinds of models. The left part is the Time Series model and the right is our approach which contains two main components: Event Extraction Module and Regression Module. Time Series Module uses time series theory to predict the FBI hate crime trends and serves as one of our base models. Event Extraction Module extracts event types and attributes, the results of which are used to constructed event-related factors. Regression Module integrates event-related factors and other predictive factors through regressive methods to predict FBI hate crime trends.

\subsection{Time Series Module}
Generally, a non-stationary time series can be transformed into a stationary one by differencing operation. Furthermore ARIMA is a typical method to model a stationary time series. A nonseasonal ARIMA model $ARIMA(p,d,q)$ has three integer parameters $p$, $d$, $q$: the order of AR, differencing, and MA, respectively. For example, a time series ${y_t}$ is said to be an $ARIMA(p,1,q)$ process if the transformation ${c_t}=\Delta {y_t}{\rm{ = }}{y_t}{\rm{-}}{y_{t-1}}=(1-L){y_t}$ follows a stationary and invertible $ARMA(p,q)$ model, where $L$ is a back-shift operator (also named lag operator). A general $ARMA(p,q)$ model is formulated as
\begin{equation} 
	{c_t} = c + \sum\limits_{i = 1}^p {{\alpha _i}} {c_{t - i}} + {\varepsilon _t} + \sum\limits_{{\rm{j}} = 1}^q {{\theta _j}{\varepsilon _{t - j}}} \label{eq1}
\end{equation}
where ${\varepsilon _t}$ is the white noise process with variance ${\sigma ^2}$, $c$ is intercept term, ${\alpha _i}$ and ${\theta _i}$ are coefficients of AR and MA processes.

Considering some financial or economic time series exhibit particular cyclical or periodic behavior, we first check whether a time series needs seasonal adjustment to remove the seasonal effect. In our study, we utilized the stats package in the R program to decompose the quarterly FBI hate crime time series ($fbi\_num$) into the seasonal, trend, and irregular components. Finally, we transform the trend component ($fbi\_num\_noseasonnal$) into a stationary time series and build an ARIMA model.

\subsection{Event Extraction Module}
\begin{figure}[htbp]
	\centering
	\includegraphics[width= 5.5 in]{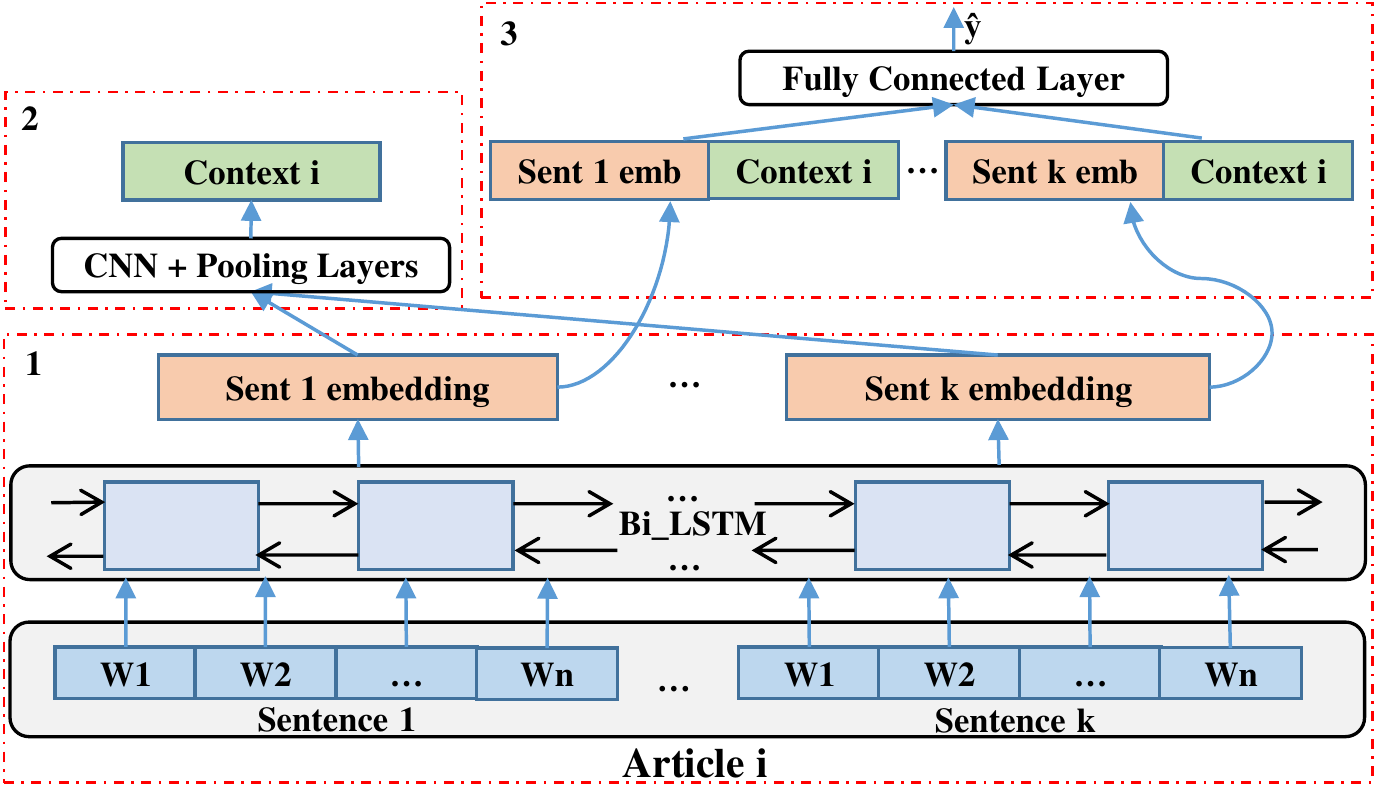}	
	\caption{Architecture of Event Extraction Module.}
	\label{fig4}	
\end{figure}

The Event Extraction Module is responsible for extracting event types and attributes, the results of which are used to construct event-related predictive factors. We follow previous work \cite{Mostafazadeh-DavaniYeh-397} and adapt Multi-Instance Learning (MIL) method developed by \cite{WangNing-405} to extract hate crime events. As shown in Figure \ref{fig4}, the event extraction module has three main essential components. 

The first component is responsible for generating sentence embeddings which are regarded as local features of an article. The BiLSTM layer is used to learn sequence information, considering the input words' past and future context information. The second component consists of a CNN layer and a Pooling layer, which scans the sentence embeddings and generates a lower-dimensional vector embedding, representing the global feature of an article. While the two previous components are responsible for generating local and global features, the third component is responsible for extracting hate crime events, including event types and attributes. Both the event types and event attributes extraction are formulated as classification tasks. The only difference between them lies that event type detection is formulated as a binary-class prediction while event attributes are treated as a multi-label classification task. 

We use the Patch Hate Crime dataset to train these two separate models and save the parameters separately.

\subsection{Regression Module}
The previous research suggests hate crime may be related to Strain Theory Criminology, Criminological Theories, or Social-Psychological Theories. We try our best to consider more alternative explainable variables than that previous literature mentioned. Due to some alternative variables are unmeasurable or severely collinear with other alternative factors, we finally left 14 variables according to data exploration(e.g., the collinear test). Furthermore, Table \ref{tab1} summarizes the alternative explainable factors related to hate crimes.

\begin{table}
	\caption{Summary of alternative explainable factors related to hate crimes.}
	\centering
	\begin{tabular}{llll}
		\toprule
		Main Stream Theories	& Alternative Factors	& Measurable	& Accepted\\
		\midrule
		\multirow{5}{1 in}{Strain Theory Criminology \cite{GreenMcfalls-406, Hall-407, Walters-408}} 
		& population	& Yes	& Yes\\
		& employment rate	& Yes	& No\\
		& unemployment rate	& Yes	& Yes\\
		& economic downturn	& Yes	& No\\
		& poverty rate	& Yes	& No\\
		\midrule
		\multirow{16}{1 in}{Criminological Theories \cite{Mills-409, MillsFreilich-410, Perry-411, KarstedtNyseth-Brehm-412}} 
		& violent crime rate	& Yes	& No\\
		& murder and nonnegligent manslaughter rate	& Yes	& Yes\\
		& rape rate	& Yes	& Yes\\
		& robbery rate	& Yes	& Yes\\
		& aggravated assault rate	& Yes	& Yes\\
		& property crime rate	& Yes	& No\\
		& burglary rate	& Yes	& Yes\\
		& larceny theft rate	& Yes	& No\\
		& homicide victims who are White	& Yes	& No\\
		& homicide victims who are Black	& Yes	& Yes\\
		& arrests violent crime	& Yes	& No\\
		& arrests weapons	& Yes	& Yes\\
		& arrests drug abuse violations	& Yes	& Yes\\
		& arrests stolen	& Yes	& No\\
		& total law enforcement employees	& Yes	& Yes\\
		& number of agencies	& Yes	& No\\
		\midrule
		\multirow{4}{1 in}{Social-Psychological Theories \cite{GreenMcfalls-406, Pettigrew-413, Agnew-414}}
		& news reports number	& Yes	& Yes\\
		& hate crime reports number	& Yes	& Yes\\
		& hate reported index	& Yes	& Yes\\
		& prejudice	& No	& No\\
		\bottomrule
	\end{tabular}
	\label{tab1}
\end{table}

Criminological strain theory \cite{Hall-415} emphasizes that the mismatching between the goals of material possessions and the means available to achieve those goals may contribute to hate crimes or violent crimes. From this view, we utilize the alternative predictors of strain theory and criminological theories left in Table \ref{tab1}. The final acceptable variables include the unemployment rate, murder and nonnegligent manslaughter rate, rape rate, robbery rate, aggravated assault rate, burglary rate, Black homicide victims, the arrested carrying or possessing weapons, the arrested abusing drug, and total law enforcement employees. We formulate the detrend $FBI\_Num\_Noseas$ prediction task as:
\begin{equation} 
\begin{array}{l}
	{\rm{fbi\_num\_noseasonnal   =  }}{\alpha _0}{\rm{ +  }}{\alpha _1}{\rm{*aggravated\_assault\_rate( - 1) }}\\
	{\rm{ +  }}{\alpha _2}{\rm{*arrests\_drug\_abuse\_violations( - 1)  +  }}{\alpha _3}{\rm{*arrests\_weapons( - 1) }}\\
	{\rm{ +  }}{\alpha _4}{\rm{*burglary\_rate( - 1)  +  }}{\alpha _5}{\rm{*homicide\_victims\_black( - 1) }}\\
	{\rm{ +  }}{\alpha _6}{\rm{*murder\_nonnegligent\_manslaughter\_rate( - 1) }}\\
	{\rm{ +  }}{\alpha _7}{\rm{*population  +  }}{\alpha _8}{\rm{*rape\_rate( - 1)  +  }}{\alpha _9}{\rm{*robbery\_rate( - 1) }}\\
	{\rm{ +  }}{\alpha _{10}}{\rm{*total\_law\_enforcement\_employees( - 1)  +  }}{\alpha _{11}}{\rm{*uner\_quar( - 1)  +  }}{\varepsilon _2}
\end{array}
\label{eq2}
\end{equation}
where ${\varepsilon _2}$ is the residual error term that represents all variation left unexplained.

The social psychological theory \cite{GreenMcfalls-406} argues that the media coverage, especially sensationalist coverage of spectacular hate crime events, contributes to hate crime contagion. We utilize event extraction technology to extract hate crime events from New York Times to construct social-psychological variables. Once the trained event extraction model predicts the New York Times news data, we obtain the detected event type list. Then we count the number of the news reports and events detected, leading to two time series: New York News articles number ($news\_num$) and Hate Crime prediction list ($event\_detected\_num$). Note that the extracted hate crime events are not the same as those mentioned in the FBI’s reports. We add these two event-related variables into formula \ref{eq2}, update it as below:
\begin{equation} 
\begin{array}{l}
	{\rm{fbi\_num\_noseasonnal   =  }}{\alpha _0}{\rm{ +  }}{\alpha _1}{\rm{*aggravated\_assault\_rate( - 1) }}\\
	{\rm{ +  }}{\alpha _2}{\rm{*arrests\_drug\_abuse\_violations( - 1)  +  }}{\alpha _3}{\rm{*arrests\_weapons( - 1) }}\\
	{\rm{ +  }}{\alpha _4}{\rm{*burglary\_rate( - 1)  +  }}{\alpha _5}{\rm{*homicide\_victims\_black( - 1) }}\\
	{\rm{ +  }}{\alpha _6}{\rm{*murder\_nonnegligent\_manslaughter\_rate( - 1) }}\\
	{\rm{ +  }}{\alpha _7}{\rm{*population  +  }}{\alpha _8}{\rm{*rape\_rate( - 1)  +  }}{\alpha _9}{\rm{*robbery\_rate( - 1) }}\\
	{\rm{ +  }}{\alpha _{10}}{\rm{*total\_law\_enforcement\_employees( - 1)  +  }}{\alpha _{11}}{\rm{*uner\_quar( - 1) }}\\
	{\rm{ +   }}{\alpha _{12}}{\rm{*event\_detected\_num  +  }}{\alpha _{13}}{\rm{*news\_num  +  }}{\varepsilon _3}
\end{array}
\label{eq3}
\end{equation}

However, the absolute amounts of $news\_num$ and $event\_detected\_num$ may not be sufficient to depict the extent that hate crime reports spread on social media at that time. We construct a relative indicator to character current hate crime transmission level, named $hate\_reported\_index$, which is calculated by:
\begin{equation} 
{\rm{hate\_reported\_index  =  }}\frac{{{\rm{event\_detected\_num}}}}{{{\rm{news\_num}}}}
\label{eq4}
\end{equation}

Finally, we integrate social-psychological related variables and criminological strain related variables into one model. Formally, we fulfill the hate crime prediction task by the following equation:
\begin{equation} 
\begin{array}{l}
	{\rm{fbi\_num\_noseasonnal   =  }}{\alpha _0}{\rm{ +  }}{\alpha _1}{\rm{*aggravated\_assault\_rate( - 1) }}\\
	{\rm{ +  }}{\alpha _2}{\rm{*arrests\_drug\_abuse\_violations( - 1)  +  }}{\alpha _3}{\rm{*arrests\_weapons( - 1) }}\\
	{\rm{ +  }}{\alpha _4}{\rm{*burglary\_rate( - 1)  +  }}{\alpha _5}{\rm{*homicide\_victims\_black( - 1) }}\\
	{\rm{ +  }}{\alpha _6}{\rm{*murder\_nonnegligent\_manslaughter\_rate( - 1) }}\\
	{\rm{ +  }}{\alpha _7}{\rm{*population  +  }}{\alpha _8}{\rm{*rape\_rate( - 1)  +  }}{\alpha _9}{\rm{*robbery\_rate( - 1) }}\\
	{\rm{ +  }}{\alpha _{10}}{\rm{*total\_law\_enforcement\_employees( - 1)  +  }}{\alpha _{11}}{\rm{*uner\_quar( - 1) }}\\
	{\rm{ +  }}{\alpha _{12}}{\rm{*event\_detected\_num  +  }}{\alpha _{13}}{\rm{*news\_num }}\\
	{\rm{ +  }}{\alpha _{14}}{\rm{*hate\_reported\_index}} + {\varepsilon _3}
\end{array}
\label{eq5}
\end{equation}

For convenience, we also name ARIMA equation \ref{eq1} Model 1, regression equation \ref{eq2} Model 2, regression equation \ref{eq3} Model 3, regression equation \ref{eq5} Model 4. We adopt Maximum Likelihood (ML) and Least Square (LS) method to estimate the parameters.

\section{Experiments}
This section reports the experimental results, including event extraction results, national hate crime prediction results (time series model and regressive models), and state-level hate crime prediction results (panel data regressive models). We adopt the data from 2007 to 2018 to fit the models, predict the hate crime trends after 2018 and compare them with the latest FBI statistics. The event extraction statistical evaluation metrics include precision, recall, and F1 score, which are calculated as follows:
\begin{equation} 
	Precision = \frac{{TP}}{{TP + FP}}
	\label{eq6}
\end{equation}
\begin{equation} 
	Recall = \frac{{TP}}{{TP + FN}}
	\label{eq7}
\end{equation}
\begin{equation}  
	\nonumber F1 = \frac{{2*Precision*Recall}}{{Precision + Recall}}= \frac{{2*TP}}{{2*TP + FP + FN}}
	\label{eq8}
\end{equation}

These performance measures provide a brief explanation of the "Confusion Metrics". True positives (TP) and true negatives (TN) are the observations that are correctly predicted. In contrast, false positives (FP) and false negatives (FN) are the values that the actual class contradicts with the predicted class.

The prediction task evaluation metrics include Adjusted R-Squared, Log-Likelihood, RMSE (Root Mean Square Error), and MAPE (Mean Absolute Percentage Error), which have been frequently used in recent research. RMSE is a mainstream accuracy indicator describing the forecasting deviation from the actual values:
\begin{equation}
	RMSE = \sqrt {\frac{1}{N}\sum\nolimits_{{\rm{t}} = 1}^N {{{({y_t} - {{\hat y}_t})}^2}} } 
	\label{eq9}
\end{equation}
Considering RMSE is sensitive to data, MAPE is employed to measure the predicted error as a percentage:
\begin{equation}
	MAPE = \frac{1}{N}\sum\nolimits_{{\rm{t}} = 1}^N {\left| {\frac{{{y_t} - {{\hat y}_t}}}{{{y_t}}}} \right|} 
	\label{eq10}
\end{equation}
where ${y_t}$ is the actual value, ${\hat y_t}$ presents the predicted value, and $N$ is the number of total observations in test set.

\subsection{Event Extraction Evaluation}
We first train and evaluate the event extraction model on the Patch Hate Crime dataset, which is divided into train set (3619), validation set (513), and test set (1039). The other parameter settings follow \cite{Mostafazadeh-DavaniYeh-397}. We repeat the experiment 10 times and report the average precision, recall, and F1 in Table \ref{tab2}. 

\begin{table}
	\caption{Event Extraction module performance on Pach Hate Crime dataset.}
	\centering
	\begin{tabular}{llll}
		\toprule
		& Precision	& Recall	& F1\\
		\midrule
		Patch Hate Crime	& 0.8162	& 0.8325	& 0.8243\\
		\bottomrule
	\end{tabular}
	\label{tab2}
\end{table}

The performance on the test set is consistent with the results reported in \cite{Mostafazadeh-DavaniYeh-397}. The trained model is saved to detect whether a New York Times news reports a hate crime event. We count the number of the news reports and events detected by the model quarterly and get two time series: New York News articles number ($news\_num$) and Hate Crime predict list ($event\_detected\_num$).

\subsection{Time Series Prediction Results}
We obtain the $fbi\_num\_noseasonnal$ time series by removing the seasonal component from the quarterly hate crime time series ($fbi\_num$). ADF test shows that it has a nonlinear trend and is not a stationary sequence. Stationary sequence $d\_fb\i_num\_noseasonal$ is obtained by differencing operation. The AR and MA orders can be identified by the Autocorrelation Function (ACF) and Partial Autocorrelation Correlation Function (PACF) test. The ACF and PACF test results suggest that the AR and MA orders are all 0. Finally, we model $fbi\_num\_noseasonnal$ with ARIMA(0,1,0), and the fitted expression is formulated as:
\begin{equation}
{\rm{fbi\_num\_noseasonnal   =  201}}{\rm{.42  +  0}}{\rm{.8776* fbi\_num\_noseasonnal( - 1)}}
\label{eq11}
\end{equation}
with Adjusted R-Squared 0.7415, log-likelihood -283.39, and Dubin-Watson statistics 1.65. We then use this fitted model to predict hate crime trends dynamically; the RMSE and MAPE are 93.11, 4.60 respectively. This model serves as one of our base models (Model 1).

\subsection{National Hate Crime Prediction Results}
We first fit Model 2 (equation \ref{eq2}), Model 3 (equation \ref{eq3}), and Model 4 (equation \ref{eq5}), then use them to forecast the hate crime values. We report the forecast evaluation metrics in Table \ref{tab3}. Each line represents a unique model, and each column is an evaluation metric. For convenient comparison, we put Model 1 evaluation together into Table \ref{tab3}. 

From Table \ref{tab3}, we can get three observations. First, regressive models outperform the ARIMA model on almost all evaluation metrics. Second, among the regressive models, the models (Model 3 and Model 4) with event-related variables outperform the ones without event-related variables (Model 2). Third, Model 4 (RMSE is 55.8065 and MAPE is 2.1355) gets the best performance among the three regressive models. It confirms what we suspect that the two directly event-extracted variables ($news\_num$ and $event\_detected\_num$) may not be sufficient, and the relative indicator $hate\_reported\_index$ we constructed is a beneficial supplementary prediction factor. The experiment results show that event extracted variables can effectively improve the hate crime prediction performance and demonstrate the effectiveness of the event extraction module in our framework.

Considering that the ARIMA model (Model 1) has the highest Adjusted R-Square, we build a model (Model 5) integrating the ARIMA model and the regressive model (Model 4) to test whether an integrated model performs best. The experimental results are also exhibited in Table 3. It shows that although the fitting evaluation metrics outperform others, it does not match for Model 4 on forecast evaluation metrics. The reason probably is that the hate crime number of quarter 1, 2019 declined precipitously. The inertia of ARIMA part contributes to more errors leading Model 5 has suboptimal evaluation on the test set.

\begin{table}
	\caption{The results of all econometric regression models.}
	\centering
	\begin{tabular}{lllll}
		\toprule
		\multirow{2}{1 in}{Models} & \multicolumn{4}{c}{Evaluation Metrics}\\
		& R-Squared	& Log Likelihood	& RMSE	& MAPE\\
		\midrule
		Model 1	& 0.7415	& -283.3861	& 93.1065	& 4.5966\\
		Model 2	& 0.7178	& -279.5405	& 62.8914	& 2.6048\\
		Model 3	& 0.7115	& -278.6714	& 86.3023	& 3.5788\\
		Model 4	& 0.7257	& -278.1902	& 55.8065	& 2.1355\\
		Model 5	& 0.7143	& -280.6130	& 57.6688	& 2.2771\\
		\bottomrule
	\end{tabular}
	\label{tab3}
\end{table}

\begin{figure}[htbp]
	\centering
	\includegraphics[width= 5.5 in]{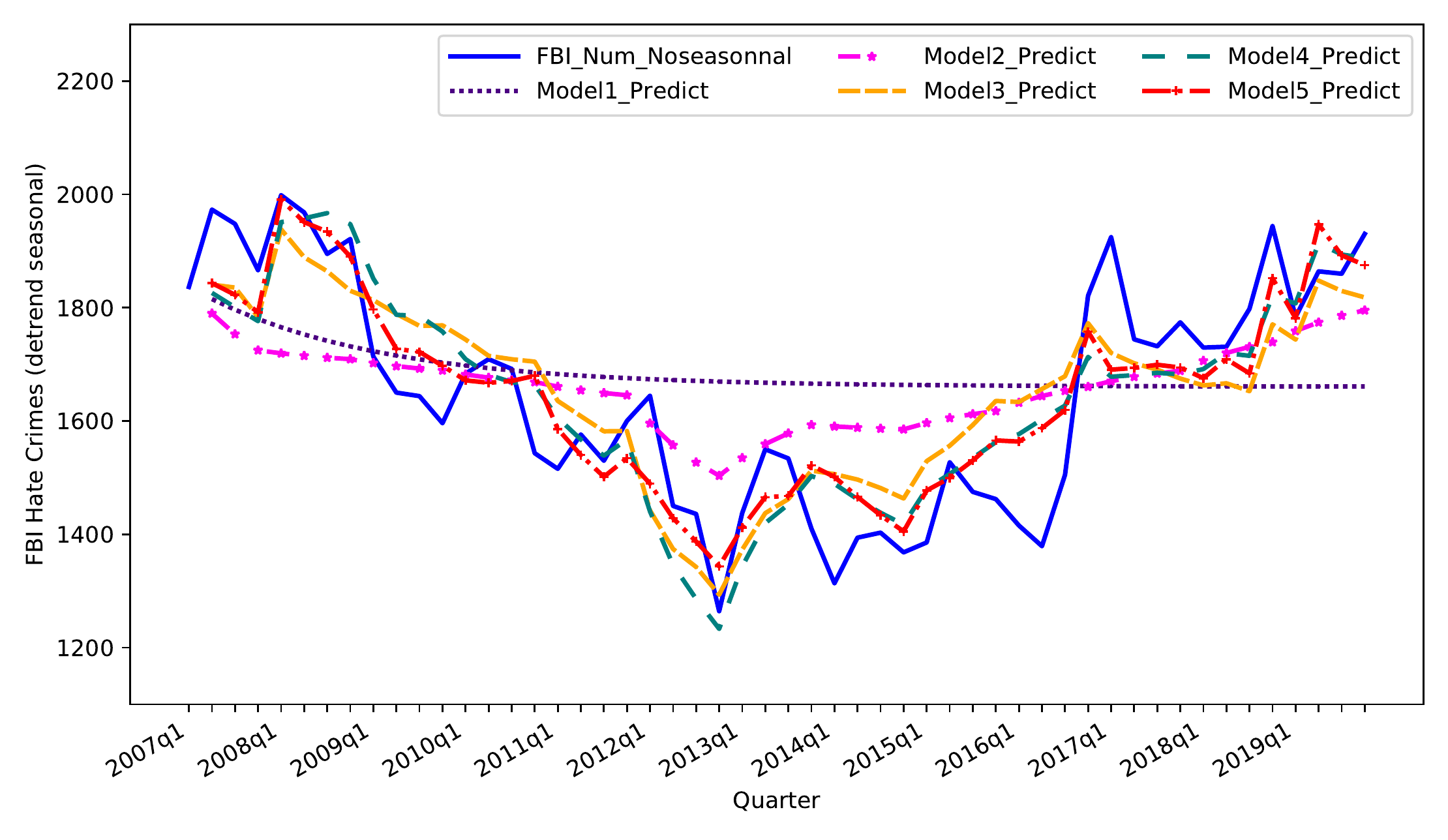}	
	\caption{The prediction performance visualization of all econometric regression models.}
	\label{fig5}	
\end{figure}

We visualize the performance by plotting the detrend hate crimes and forecast values predicted in Figure \ref{fig5}. It straightforwardly shows that the models considering the event extracted factors perform better than those without them.

However, we note that the gap between the predicted hate crime numbers and the actual values is a bit large between late 2016 and 2017. We find that the "Trump Effect" could well explain this phenomenon. According to Stephen and Griffin's research \cite{EdwardsRushin-416}, they concluded that it was not just Trump's inflammatory rhetoric throughout the political campaign from November of 2016 that caused hate crimes to increase. Rather, they also argued that Trump's subsequent election as President of the USA validated that rhetoric in perpetrators' eyes and fueled the hate crime surge.

\subsection{State-level Hate Crime Prediction Results}
Our approach can be easily adapted to predict state-level hate crime trends. We utilize the panel data analysis method to fulfill this task. The process includes five main steps. a). Detect which state is involved in the news reporting hate crimes. b). Prepare data for panel data regressive analysis. c). Perform Fix/Random effects test to determine the proper form of panel data equations. d). Estimate the parameters and forecast. e). Forecast state-level hate crime trends.

We first perform Named Entity Recognition (NER) to identify which states are involved in the hate crimes. To enhance recognition accuracy, we collect 730000 American place or institute names and map these names to their corresponding states. We sample 500 news and manually annotate the states where the hate crimes take place. Two annotators and our state recognization program achieved a 0.79 agreement on Cohen’s Kappa. This favorable performance promises that our state recognition method is qualified for preparing reliable panel data variables.

Unlike the previous method mentioned in the national prediction task, we count the number of news reports and events detected by the model quarterly and group them by state. Hence, each state gets two time series: New York News articles number involved with this state ($*\_news\_num$) and Hate Crime predict list ($*\_event\_detected\_num$), where the asterisk wildcard * denotes any state. Event-related factors and other factors are prepared to panel data form.

Considering the sparsity of FBI hate crime data and event extraction results by state, we removed four states whose data is incomplete. There are 47 states left with a hate crime proportion of 99.4\%. We perform panel data regression analysis on this panel data. Then Hausman Test is performed to test Fix or Random effects. The test results suggest that adopting the equation with Fix effects is a better choice. 

We use LS to estimate the parameters using the data between 2007 quarter 1 and 2018 quarter 4. Like the national hate crime prediction, we forecast the hate crimes between 2019 quarter 1 and 2019 quarter 4. The experiment results of Model 6 (with no event-extracted variables) and Model 7 (with event-extracted variables) are exhibited in Table \ref{tab4}. The result shows that event-related variables help Model 7 surpass Model 6 on RMSE 14.9 percent and MAPE 2.7 percent. 

\begin{table}
	\caption{The results of panel data regressive models.}
	\centering
	\begin{tabular}{lllll}
		\toprule
		\multirow{2}{1 in}{Models} & \multicolumn{4}{c}{Evaluation Metrics}\\
		& R-Squared	& Log Likelihood	& RMSE	& MAPE\\
		\midrule
		Model 6	& 0.6177	& -4844.1370	& 16.7421	& 39.8546\\
		Model 7	& 0.6222	& -4800.1699	& 14.2317	& 37.1102\\
		\bottomrule
	\end{tabular}
	\label{tab4}
\end{table}

We perform the Levene Test for equal variances. Levene statistic is 9.49e-06, and the p-value is 0.9975, showing that the two prediction series have the same variance. Paired Samples t-Test statistic is -7.931, and the p-value is 2.089e-13, showing that the two prediction series do not have identical average (expected) values. The mean of actual sampling values on test data is 38.6956, whereas that of Model 6 is 34.7538, and that of Model 7 is 35.1390. The results of Paired Samples Test show that Model 7 significantly outperforms Model 6. 

The results (including RMSE, MAPE, and Paired Samples Test) demonstrate the effectiveness and generalization of the event extraction module of our framework. It could not only be used to predict national hate crime trends but also state-level trends.

\subsection{Shortcomings and Future Work}
Existing literature on causes of hate crime may include prejudice theory, criminological theory, social-psychological theory. Unfortunately, there is limited research on the general reason causing a hate crime. One shortcoming of this study is that we only introduce some criminological and social-psychological variables as our predictive factors. For convenient comparison, we focused on demonstrating event-extraction-related factors could help enhance the prediction ability, thus ignored many factors of other theories, e.g., prejudice theory. We will try to consider more factors introduced in other literature research and verify their effectiveness in the future.

In future work, we will collect more data and do breakpoint tests to verify the "Trump effect", encode this variable into a two-stage model, and check whether it would enhance the model performance. We are also continuing to design some new models, including Copula-based models and neural network models.

\section{Conclusion}
Hate crimes continue to be a pervasive problem in the United States. The FBI's UCR Program collects national hate crime data and publishes statistics reports every year, but the statistics are mainly released more than one year later. It is important for practitioners, policy-makers, researchers to grasp the national and local hate crime trends in time. We propose a framework by introducing NLP technologies, especially event extraction technologies, into national hate crime trend prediction to alleviate the challenge. This framework also can be easily adapted to predict state-level hate crime trends by integrating panel data analysis. The experiment results show that event extracted factors significantly enhance the prediction performance. Our framework broadens the approaches of hate crime trend prediction. It can help provide valuable holistic and strategic insight for law enforcement agencies and justify specific legislation in advance for lawmakers.

\bibliographystyle{unsrt}  
\bibliography{references}

\end{document}